\documentclass[letterpaper, 10 pt, conference]{ieeeconf}  

\IEEEoverridecommandlockouts                              

\makeatletter

\let\NAT@parse\undefined

\makeatother

\usepackage{amssymb}
\usepackage{amsmath}
\usepackage{xspace}
\usepackage{physics}
\usepackage{graphicx}
\usepackage{multirow}
\usepackage{siunitx}
\usepackage{pifont}
\usepackage{subcaption}
\usepackage{array}
\usepackage{hyperref}

\newcommand{\figref}[1]{Figure~\ref{#1}} 

\newcommand{\tabref}[1]{Table~\ref{#1}} 

\newcommand{\secref}[1]{Section~\ref{#1}}

\makeatletter

\DeclareRobustCommand\onedot{\futurelet\@let@token\@onedot}

\newcommand\@onedot{\ifx\@let@token.\else.\null\fi\xspace} 

\newcommand\eg{\emph{e.g}\onedot} 

\newcommand\ie{\emph{i.e}\onedot} 

\newcommand\cf{\emph{cf}\onedot} 

\newcommand\wrt{w.r.t\onedot} 

\newcommand\etal{\emph{et al}\@onedot} 

\makeatother

\DeclareMathOperator*{\argmax}{arg\,max}

\newcommand*{\tran}{^{\mkern-1.5mu\mathsf{T}}}

\title{\LARGE \bf
Continuous Self-Localization on Aerial Images Using Visual and Lidar Sensors
}

\author{Florian Fervers$^{1}$, Sebastian Bullinger$^{1}$, Christoph Bodensteiner$^{1}$, Michael Arens$^1$, Rainer Stiefelhagen$^2$
\thanks{$^{1}$Fraunhofer IOSB, Ettlingen, Germany {\tt\small <firstname>. <lastname>@iosb.fraunhofer.de}}%
\thanks{$^{2}$Karlsruhe Institute of Technology, Karlsruhe, Germany
        {\tt\small rainer.stiefelhagen@kit.edu}}%
}

\begin{document}

\maketitle
\thispagestyle{empty}
\pagestyle{empty}

\begin{abstract}

This paper proposes a novel method for \mbox{geo-tracking}, \ie continuous metric self-localization in outdoor environments by registering a vehicle's sensor information with aerial imagery of an unseen target region. Geo-tracking methods offer the potential to supplant noisy signals from global navigation satellite systems (GNSS) and expensive and hard to maintain prior maps that are typically used for this purpose. The proposed geo-tracking method aligns data from on-board cameras and lidar sensors with geo-registered orthophotos to continuously localize a vehicle. We train a model in a metric learning setting to extract visual features from ground and aerial images. The ground features are projected into a \mbox{top-down} perspective via the lidar points and are matched with the aerial features to determine the relative pose between vehicle and orthophoto.

Our method is the first to utilize on-board cameras in an \mbox{end-to-end} differentiable model for metric self-localization on unseen orthophotos. It exhibits strong generalization, is robust to changes in the environment and requires only geo-poses as ground truth. We evaluate our approach on the KITTI-360 dataset and achieve a mean absolute position error (APE) of 0.94m. We further compare with previous approaches on the KITTI odometry dataset and achieve \mbox{state-of-the-art} results on the geo-tracking task. \footnote[3]{Project page: \url{https://fferflo.github.io/projects/contselfloc22}}

\end{abstract}

\section{INTRODUCTION}

An accurate method for self-localization is essential for ground-vehicle navigation. This is the case for human drivers supported by navigational systems as well as fully autonomous driving suites. Short-term trajectories can be estimated accurately via odometry methods utilizing visual, lidar or inertial measurements. However, even under ideal benchmark conditions, state-of-the-art odometry methods suffer from a drift upwards of 5m after traveling for 1km \cite{kitti-benchmark}. GNSS can be used to alleviate long-term drift, but suffer from outages and noisy measurements (\eg due to the multipath effect \cite{karaim2018gnss}). A common alternative to GNSS is to continuously realign the local trajectory with a pre-built map of the environment \cite{bresson2017simultaneous}. While ground-based maps (\eg using lidar point clouds) are costly to produce and to keep up-to-date, two-dimensional grid-maps such as aerial orthophotos are available globally and offer the potential for self-localization in unseen areas without corresponding ground maps.

Recent work in this area can be roughly divided into two categories: (1)~Geo-localization methods use no or only a coarse (\eg city-scale) initial pose estimate and determine the location by matching the ground vehicle data against a database of geo-registered aerial images. (2)~\mbox{Geo-tracking} methods start from a known pose and continuously track the vehicle's movement on aerial images or semantic maps provided by geographic information systems (GIS). While state-of-the-art geo-localization approaches \cite{shi2019spatial,rodrigues2021these,wang2021each} have high recall on benchmark datasets \cite{workman2015localize,liu2019lending} they suffer from low metric accuracy (\eg when employed in a Markov localization scheme \cite{heng2019project}). Further, initial pose estimates are often known or can be provided by cheap GNSS receivers.

Geo-tracking methods are applicable when the initial pose is known, and have demonstrated feasible results using input from visual cameras \cite{senlet2011framework}, range scanners \cite{veronese2015re,vora2020aerial} and both \cite{miller2021any}. Nevertheless, there are several limitations of current approaches to geo-tracking that we address in this work:

\begin{itemize}
	\item Several methods \cite{miller2021any,tang2020rsl,tang2021self,Tang2021GetTT} train and test on data from the same city area. Others \cite{chu2015accurate,zha2021map} assume that the ground region or aerial data to be tested on has already been seen during the training stage. We emphasize that one of the advantages of using orthophotos or GIS for self-localization is not having to travel through the target area beforehand to gather training data or build ground maps. We train and test our method on entirely different datasets from the United States and Germany to demonstrate the strong generalization of our model to unseen ground and aerial data.
	\item Some methods rely on specific environment features such as edges of buildings \cite{vysotska2017improving,kim2019fusing,tang2020rsl,tang2021self,Tang2021GetTT} or semantic features \cite{miller2021any,yan2019global}. This limits the applicability to locations where these features are present and visible, \ie urban regions.
	\item Several related works \cite{senlet2011framework,noda2010vehicle,veronese2015re,vora2020aerial} rely on matching low-level visual or lidar-reflectance features and assume that ground and aerial images have a similar low-level appearance. However, dynamic objects in the ground and aerial data do not match unless aerial and ground data are captured synchronously. Further, aerial images from sources such as Google Maps \cite{googlemaps} or Bing Maps \cite{bingmaps} are captured over several years \cite{lesiv2018characterizing} and potentially miss recent changes (\eg from construction sites and vegetation growth).
\end{itemize}

We present a new method for geo-tracking that aligns ground with aerial images via learned visual features. We use lidar point clouds captured synchronously with the ground data to project features extracted from ground images into a nadir (\ie top-down) perspective. The matching provides a two-dimensional relative pose \wrt the aerial image that translates into latitude, longitude and bearing of the vehicle using the geo-registered information of the aerial image.

Our method is aimed to complement existing odometry and simultaneous mapping and localization (SLAM) methods by providing continuous geo-registration that reduces long-term drift to within the registration error. We choose a simple inertial odometry method based purely on measurements of acceleration and angular velocity and show that in combination with our geo-registration it achieves meter-accurate global poses regardless of sequence length.

The main contribution of this work is a novel method for continuous geo-tracking on orthophotos with meter-level accuracy that exhibits strong generalization, does not require hand-crafted features or ground truth and is robust to changes in the environment. Our experiments demonstrate that we achieve state-of-the-art geo-tracking results on the KITTI odometry dataset \cite{Geiger2013IJRR}.

\section{RELATED WORKS}

\begin{table*}[h]
	\vspace{2mm}
	\caption{Related Works. (a) Handcrafted and Semantic Features. (b) Low-level Features. (c) End-to-end Learnable Features.}
	\label{tab:related_works}
	\hspace{-2mm}\begin{tabular}{p{1mm} p{26.5mm}|c|c|c|c|>{\centering\arraybackslash}p{32mm}}
		& Authors & Ground data & Aerial data & Extracted features & Tracking & Evaluated on\\
		\cline{2-7}
		\rule{0pt}{2.2ex}\multirow{11}{*}{(a)} & Vysotska \etal \cite{vysotska2017improving} & Lidar & OpenStreetMaps & Buildings & Pose graph & Non-public data\\
		& Kim \etal \cite{kim2019fusing} & Lidar & Orthophoto & Buildings & Particle filter & CULD \cite{jeong2018complex} \\
		& K{\"u}mmerle \etal \cite{kummerle2011large} & Lidar or camera & Orthophoto & Vertical structures & Particle filter & Non-public data\\ 
		& Wang \etal \cite{wang2017flag} & Camera & Orthophoto & Vertical structures & Particle filter & Non-public data \\
		& Pink \etal \cite{pink2008visual} & Camera & Orthophoto & Lane markings & - & Non-public data\\
		& Javanmardi \etal \cite{javanmardi2017towards} & Lidar & Orthophoto & Lane markings &  - & Non-public data\\
		& Viswanathan \etal \cite{viswanathan2016vision} & Lidar and camera & Orthophoto & Ground-nonground & Particle filter & Non-public data \\
		& Brubaker \etal \cite{brubaker2013lost} & Camera & OpenStreetMaps & Trajectory & Custom Bayesian filter & KITTI \cite{Geiger2013IJRR} \\
		& Floros \etal \cite{floros2013openstreetslam} & Camera & OpenStreetMaps & Trajectory & Particle filter & KITTI \cite{Geiger2013IJRR}\\
		& Yan \etal \cite{yan2019global} & Lidar & OpenStreetMaps & Semantic descriptor & Particle filter & KITTI \cite{Geiger2013IJRR}\\
		& Miller \etal \cite{miller2021any} & Lidar and camera & Orthophoto & Semantic classes & Particle filter & KITTI \cite{Geiger2013IJRR}\\
		\cline{2-7}
		\rule{0pt}{2.2ex}\multirow{4}{*}{(b)} & Veronese \etal \cite{veronese2015re} & Lidar & Orthophoto & - & Particle filter & Non-public data \\
		& Vora \etal \cite{vora2020aerial} & Lidar & Orthophoto & - & Extended Kalman Filter & Non-public data \\
		& Senlet \etal \cite{senlet2011framework} & Camera & Orthophoto & Edges & Particle filter & Non-public data\\
		& Noda \etal \cite{noda2010vehicle} & Camera & Orthophoto & SURF & - & Non-public data\\
		\cline{2-7}
		\rule{0pt}{2.2ex}\multirow{4}{*}{(c)} & Tang \etal \cite{tang2020rsl,tang2021self,Tang2021GetTT} & Radar or lidar & Orthophoto & Learned & - & Robotcar \cite{barnes2020oxford}, KITTI \cite{Geiger2013IJRR}, Mulran \cite{kim2020mulran}\\
		& Zhu \etal \cite{zhu2020agcv} & Lidar & Orthophoto & Vertical structures & Pose graph & KITTI \cite{Geiger2013IJRR} \\
		& \textbf{Ours} & Lidar and camera & Orthophoto & Learned & Extended Kalman Filter & \hspace{-0.5mm}KITTI \cite{Geiger2013IJRR}, KITTI-360 \cite{liao2021kitti}
	\end{tabular}
	\vspace{-5mm}
\end{table*}
\renewcommand{\arraystretch}{1.0}

\subsection{Handcrafted and Semantic Features}
\label{sec:related-works-handcrafted}

Most geo-tracking approaches involve identifying abstract hand-crafted or semantic features in both ground and aerial data that are matched to find the vehicle's relative pose (\cf \tabref{tab:related_works}a). The ground data is captured by visual cameras and/or range scanners (\ie lidar, radar), and the aerial data is represented by orthophotos or grid-maps from GIS such as OpenStreetMaps (OSM) \cite{OpenStreetMap}. Different types of features have been used for this purpose in previous works, such as building outlines \cite{vysotska2017improving,kim2019fusing}, general vertical structures \cite{kummerle2011large,wang2017flag}, lane markings \cite{pink2008visual,javanmardi2017towards}, a binary ground-nonground distinction \cite{viswanathan2016vision}, the driven trajectory \cite{brubaker2013lost,floros2013openstreetslam} and multiple semantic classes \cite{yan2019global,miller2021any}.

Semantic and handcrafted features are typically defined to be invariant to seasonal and daylight variations, but also discard much of the information contained in the input data that could otherwise be leveraged for the tracking task. This limits the applicability to areas where the desired type of feature is abundant, \ie mostly urban regions with a sufficient amount of buildings.

\subsection{Low-level Features}
\label{sec:related-works-low-level}

In contrast to hand-crafted or semantic features, methods based on low-level features allow leveraging the input data on a more fine-grained level by directly comparing color and reflectance values \cite{veronese2015re,vora2020aerial} or low-level visual features \cite{senlet2011framework,noda2010vehicle} (\cf \tabref{tab:related_works}b). The camera or lidar data is first projected into a top-down perspective before applying the matching algorithm.

Methods based on low-level features are negatively impacted by changes in illumination and material appearance when ground and aerial images are taken years apart. Furthermore, the different surface areas of vertical structures are typically only visible from either the aerial view (\eg roofs) or the ground view (\eg walls) and can thus not be exploited for the tracking task. As a consequence, low-level feature methods rely almost entirely on contrast between bright lane markings, sidewalks and darker roads.

\subsection{End-to-end Learnable Features}
\label{sec:related-works-end-to-end}

End-to-end trainable models for the geo-tracking task have only recently gained some attention in the research community (\cf \tabref{tab:related_works}c). Unlike low-level features, learned features allow exploiting vertical structures in the environment and can be trained to be invariant to illumination and appearance changes due to a higher level of abstraction. Unlike handcrafted and semantic features, these models do not blanket discard part of the input data, but instead learn to predict invariant features in a data-centric manner. \mbox{End-to-end} learned features can thus combine the advantages of both semantic or handcrafted features and low-level features.

Previous methods based on end-to-end trainable models \cite{tang2020rsl,tang2021self,zhu2020agcv,Tang2021GetTT} utilize only input from range-scanners (\ie radar and lidar) on the ground vehicle. The works by Tang \etal \cite{tang2020rsl,tang2021self,Tang2021GetTT} also discard lidar points on the ground plane and thereby focus entirely on vertical structures.

\subsection{Cross-View Geo-localization}

Cross-view geo-localization refers to the problem of matching a given ground image against a dataset of geo-registered aerial images to find the geo-location of the vehicle. Current state-of-the-art approaches \cite{shi2019spatial,rodrigues2021these,wang2021each} achieve high recall on corresponding benchmark datasets like CVUSA \cite{workman2015localize} and CVACT \cite{liu2019lending}. Such models can also be used to perform coarse geo-tracking (\eg in combination with particle filters \cite{heng2019project}), but do not compete with the metric accuracy of dedicated geo-tracking approaches.

\subsection{Placement of Our Work}

We propose an end-to-end trainable model for geo-tracking on aerial images in unseen areas. Our work is the first to utilize ground cameras in an end-to-end differentiable model since previous similar works have only used range scanners for this purpose (\cf \secref{sec:related-works-end-to-end}). Camera images both contain rich visual information and have a smaller domain gap to the aerial imagery than range scans.

The model extracts high-level features from aerial and ground images which are more robust to appearance changes than low-level features (\cf \secref{sec:related-works-low-level}) and can exploit vertical structures. In contrast to handcrafted and semantic high-level features (\cf \secref{sec:related-works-handcrafted}), the model does not focus on a predefined subset of the input data and thereby discard information that can possibly help improve tracking performance.

\section{METHOD}

Our method extracts visual features from both ground and aerial images and aligns them from a top-down perspective to find the two-dimensional pose of the vehicle relative to the aerial image. We train a model in a metric learning setting to predict similar features for matching pixels of aerial and ground images and dissimilar features for non-matching pixels. The transformation between projected ground features and aerial features that maximizes the sum of pixel matching scores defines the vehicle's pose relative to the aerial image.

\subsection{Input and Coordinate Systems}
\label{sec:coordinate_systems}

We use a scaled Spherical Mercator projection \cite{mercator} to transform geo-poses into a locally euclidean, metric coordinate system of the earth's surface. All poses are represented by 3-DoF transformations $T = [R|t] \in \text{SE}(2)$ between local vehicle coordinates and global coordinates. The global coordinate system is centered on the initial pose estimate and stores coordinates in east-north order.

To perform the geo-registration of the vehicle at a given point in time, the method takes the following inputs:
\begin{enumerate}
	\item An initial pose estimate $T_A = [R_A|t_A] \in \text{SE}(2)$ and an aerial image $I_A \in \mathbb{R}^{h_A \times w_A \times 3}$ centered on $t_A$ and rotated by $R_A$ with a resolution of $q \in \mathbb{R}$ meters per pixel and dimensions $h_A \times w_A$.
	\item Multiple ground images $I_{G_i} \in \mathbb{R}^{h_{G_i} \times w_{G_i} \times 3}$ captured synchronously by $n_G$ different cameras on the vehicle with dimensions $h_{G_i} \times w_{G_i}$ and $i \in \{0, ..., n_G - 1\}$.
	\item A lidar point cloud $L \in \mathbb{R}^{n_L \times 3}$ with $n_L \in \mathbb{N}$ three-dimensional points in vehicle coordinates captured by one or more lidar scanners.
\end{enumerate}

All data from the ground vehicle are acquired synchronously up to a negligible delay from a pose $T_G \in \text{SE}(2)$ and represent the measured state of the environment at the given point in time. Our registration method estimates the relative transformation $T_{G \rightarrow A}$ between prior pose estimate $T_A$ and vehicle pose $T_G$.

\begin{figure*}
	\centering
	\vspace{2.5mm}
	\includegraphics[width=160mm]{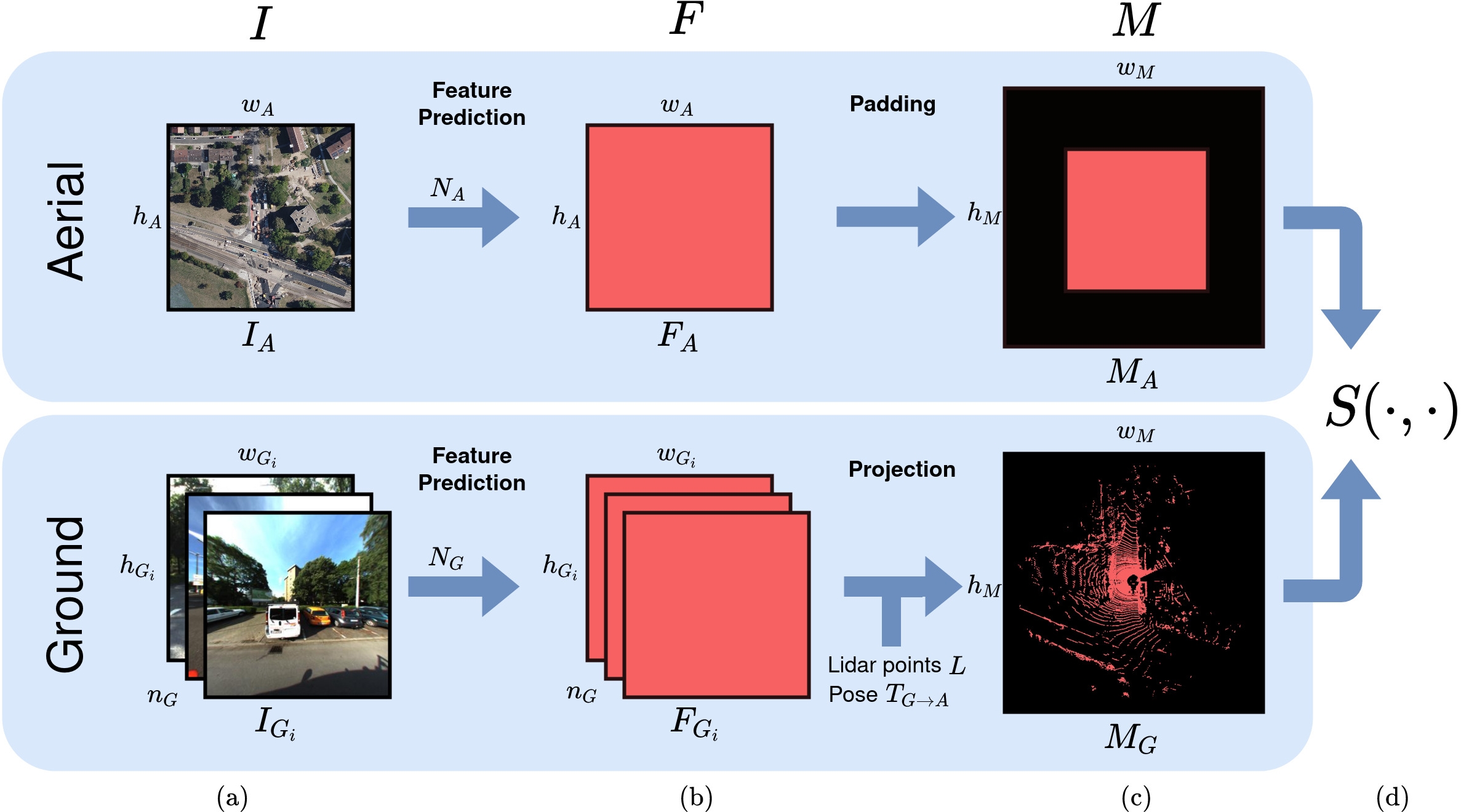}
	\caption{Summary of the steps required to compute the likelihood of a relative pose hypothesis $T_{G \rightarrow A}$ for the ground vehicle. (a) The input consists of an aerial image $I_A$ centered and rotated to match the prior pose estimate $T_A$, and $n_G$ images $I_{G_i}$ captured synchronously by cameras on the ground vehicle, as well as a lidar point cloud $L$. (b) The aerial and ground images are processed by separate aerial and ground networks $N_A$ and $N_G$ to predict pixel-wise feature maps $F_A$ and $F_{G_i}$. (c) All ground image features are projected onto a single top-down grid-map $M_G$ using the lidar point cloud $L$ and pose hypothesis $T_{G \rightarrow A}$. The aerial features are padded to produce the aerial grid-map $M_A$ that matches the size of $M_G$. Red pixels in $M_A$ and $M_G$ refer to valid features, black pixels to missing features. (d) The grid-maps are compared using the similarity function $S$ to produce a confidence score for the pose hypothesis. Multiple hypotheses are tested jointly by computing $S$ using a cross-correlation.\vspace{-5mm}}
	\label{fig:summary}
\end{figure*}

\subsection{Feature Map Computation}

We consider different hypotheses $h \in \mathcal{H}$ for the vehicle pose $T_{G^{(h)}}$. For each candidate $h$ the ground image features of all cameras are first transformed from the vehicle frame $T_{G^{(h)}}$ into $T_A$ by the transformation $T_{G^{(h)} \rightarrow A}$ and then projected via the lidar points onto a top-down grid-map $M_G^{(h)} \in \mathbb{R}^{h_M \times w_M \times c}$. The grid-map $M_G^{(h)}$ for hypothesis $h$ shall match the aerial grid-map $M_A \in \mathbb{R}^{h_M \times w_M \times c}$ if and only if $h$ is the correct hypothesis, \ie $T_{G^{(h)}} = T_{G}$. The maps have spatial dimensions $h_M \times w_M$ and $c$ channels. The true pose hypothesis $\hat{h}$ is then determined as
\begin{equation}
	\label{eq:argmax}
	\hat{h} = \argmax_{h \in \mathcal{H}} S(M_A, M_G^{(h)})
\end{equation}
where $S(\cdot, \cdot)$ measures the similarity between the two grid-maps. This is akin to a maximum likelihood estimation, but uses an uncalibrated likelihood $S$. We train our model to output high values for $S$ if $h$ is the correct hypothesis, and low values for incorrect hypotheses. A summary of the system is shown in \figref{fig:summary}.

The measured lidar points are not used as an input to compute the feature vectors themselves, but rather only serve to project features from ground view into a top-down perspective.

\subsubsection{Feature prediction}

To compute $M_A$ and $M_G^{(h)}$ for a given pose hypothesis $h$, the aerial and ground images are first processed by separate aerial and ground encoder-decoder networks $N_A$ and $N_G$. This produces pixel-level feature maps with the original image resolutions and $c$ channels (\cf \figref{fig:summary}b).
\begin{align}
	F_A &= N_A(I_A) \in \mathbb{R}^{h_A \times w_A \times c} \\
	F_{G_i} &= N_{G}(I_{G_i}) \in \mathbb{R}^{h_{G_i} \times w_{G_i} \times c} \text{   for } i \in [0..n_G - 1]
\end{align}

\subsubsection{Feature projection}

The visual features are transferred from the ground feature maps $F_{G_i}$ onto the lidar points $L$ by projecting the points onto all camera planes and sampling the feature maps at the projected pixel locations. When multiple cameras observe the same lidar point due to an overlapping field of view, the corresponding pixel features are mean-pooled to produce a single feature vector for the lidar point. This yields a feature matrix $F_L \in \mathbb{R}^{n_L \times c}$ for the $n_L$ lidar points.

Next, the features stored in $F_L$ are transformed into the top-down grid-map $M_G^{(h)} \in \mathbb{R}^{h_M \times w_M \times c}$ that matches the pixel resolution $q$ of the aerial image. The lidar points are first transformed by $T_{G^{(h)} \rightarrow A}=[R^{(h)}|t^{(h)}]$ into the coordinate system $T_A$ and then projected onto two-dimensional pixels in $M_G^{(h)}$ using an orthogonal projection along the $z$-axis. The features of lidar points that are projected onto the same pixel in $M_G^{(h)}$ are max-pooled to produce a single feature vector. Since lidar points generally do not cover all grid-map pixels, $M_G^{(h)}$ represents a sparse pixel-level feature map (\cf \figref{fig:summary}c).

Lastly, the aerial network output $F_A$ is symmetrically padded with zeros in both spatial dimensions to produce the aerial grid-map $M_A$ with spatial dimensions $h_M \times w_M$ that is used for matching with $M_G^{(h)}$ (\cf \figref{fig:summary}c).


\subsection{Feature Map Alignment}\label{sec:feature_map_alignment}

Given the two feature maps $M_A$ and $M_G^{(h)}$ with equal dimensions, the function $S(\cdot, \cdot)$ computes their matching score as the average of the similarities $s(\cdot, \cdot)$ of individual pixel embeddings. $S$ only considers pixels where both $M_A$ and $M_G^{(h)}$ store valid features, \ie it ignores padded pixels in $M_A$ and pixels in $M_G^{(h)}$ without projected lidar points. We define the set of valid pixels for a hypothesis $h$ as $\mathcal{X}^{(h)} \subset \mathbb{N}^2$.

The matching score between individual pixel embeddings $a \in \mathbb{R}^c$ and $g \in \mathbb{R}^c$ is measured as their cosine similarity:
\begin{equation}
	\label{eq:pixel-distance}
	s(a, g) = \frac{a}{\norm{a}} \cdot \frac{g}{\norm{g}} \in [-1, 1]
\end{equation}

The similarity function $S$ is defined as the average similarity of the features stored per valid pixel and measures the confidence assigned to the hypothesis $h$ as shown in \eqref{eq:similarity-score}.
\begin{equation}
	\label{eq:similarity-score}
S(M_A, M_G^{(h)}) = \frac{1}{|\mathcal{X}^{(h)}|} \sum_{(x, y) \in \mathcal{X}^{(h)}} s(M_A(x, y), M_G^{(h)}(x, y))
\end{equation}

We further consider only hypotheses where the proportion of valid ground features that are matched with aerial features is above a threshold parameter $\alpha$. This rules out hypotheses with no or only a small overlap between $M_A$ and $M_G^{(h)}$ (\ie when $\norm{t^{(h)}}$ is too large). In this case, the measurement of $S(M_A, M_G^{(h)})$ is unreliable.

To determine the maximum likelihood hypothesis according to \eqref{eq:argmax}, the expression $S(M_A, M_G^{(h)})$ has to be evaluated for all hypotheses $h \in \mathcal{H}$. This becomes unfeasible for large $|\mathcal{H}|$. To alleviate this problem, we choose the set of hypotheses as a pixel-spaced grid of translations around the origin. This allows the corresponding confidence scores to be computed jointly via cross-correlation which reduces to a simple element-wise multiplication in the Fourier domain \cite{barnes2019masking}. The step is repeated for a discrete set of possible rotations $\mathcal{H}_R \subset \text{SO}(2)$ that are chosen as a hyperparameter.

The result of the matching step are the confidence scores $s^{(h)} = S(M_A, M_G^{(h)}) \in [-1, 1]$ for all pose hypotheses \mbox{$h \in \mathcal{H}$}.

\subsection{Training and Loss}

In the previous sections, we presented a model that predicts the confidence scores of a set of pose hypotheses. The entire model including projections and transformations into the Fourier domain is end-to-end differentiable up to the confidence scores.

Each training sample defines one positive hypothesis $h_p$ based on the ground truth pose of the vehicle, and multiple negative hypotheses $h_n \in \mathcal{H}_n$ with $\mathcal{H} = \mathcal{H}_n \cup \{ h_p\}$. We employ a soft triplet loss on each pair of positive and negative hypotheses and a soft \textit{online hard example mining} (OHEM) \cite{shrivastava2016training} strategy to train the model to predict discriminative pixel embeddings.

A triplet is defined by the anchor $M_A$, the positive example $M_G^{(h_p)}$ and the negative example $M_G^{(h_n)}$. The classical hard triplet loss aims to minimize the distance of the positive example to the anchor and maximize the distance of the negative example to the anchor by at least a margin $m$ (see \eqref{eq:triplet-hard}). The distance is represented by the negative similarity $d_h := -S(M_A, M_G^{(h)})$ between anchor and example.
\begin{equation}
	\label{eq:triplet-hard}
	l_{\text{hard}}(h_n) = \max(\Delta(h_n), 0)
\end{equation}
\begin{equation*}
	\text{with  } \Delta(h_n) = d_{h_p} + m - d_{h_n}
\end{equation*}

Taking the average of $l_{\text{hard}}(h_n)$ for all possible $h_n \in \mathcal{H}_n$ results in very small loss values as soon as the majority of triplets are classified correctly. This prevents the model from learning from hard negative hypotheses with \mbox{$\Delta(h_n) > 0$}. Instead, we employ an OHEM strategy during training by only considering hard negative hypotheses and taking the average of $l_{\text{hard}}(h_n)$ for all remaining hypotheses with $\Delta(h_n) > 0$. The loss function in \eqref{eq:hard_loss} encodes a hard triplet loss with an OHEM strategy.
\vspace{-2mm}\begin{equation}
	\label{eq:hard_loss}
	\mathcal{L}_{\text{hard}} = \frac{1}{|\mathcal{H}'_n|}\sum_{h_n \in \mathcal{H}'_n} l_{\text{hard}}(h_n) = \frac{\displaystyle \sum_{h_n \in \mathcal{H}_n} l_{\text{hard}}(h_n)}{\displaystyle \sum_{h_n \in \mathcal{H}_n} \dv{l_{\text{hard}}}{\Delta} \left(h_n\right)}
\end{equation}
\begin{equation*}
	\text{with  } \mathcal{H}'_n = \{h_n \in \mathcal{H}_n | \Delta(h_n) > 0 \}
\end{equation*}
The derivative $\dv*{l_{\text{hard}}}{\Delta}$ in \eqref{eq:hard_loss} acts as a binary indicator for a hard negative triplet with $\Delta(h_n) > 0$.

To improve the stability of the training, we smoothly approximate $\mathcal{L}_{\text{hard}}$ with a soft loss function $\mathcal{L}_{\text{soft}}$. To this end, we first replace the relu function $l_{\text{hard}}$ with a smooth softplus function as shown in \eqref{eq:triplet-soft}.
\begin{equation}
	\label{eq:triplet-soft}
	l_{\text{soft}}(h_n) = \delta \ln (1 + \exp \frac{\Delta(h_n)}{\delta})
\end{equation}

The temperature parameter $\delta$ determines the hardness of $l_{\text{soft}}$. We define the new smooth loss function $\mathcal{L}_{\text{soft}}$ as follows:
\begin{equation}
	\label{eq:soft_loss}
	\mathcal{L}_{\text{soft}} = \frac{\displaystyle \sum_{h_n \in \mathcal{H}_n} l_{\text{soft}}(h_n)}{\displaystyle \sum_{h_n \in \mathcal{H}_n} \dv{l_{\text{soft}}}{\Delta}(h_n)}
\end{equation}
The derivative $\dv*{l_{\text{soft}}}{\Delta}$ in \eqref{eq:soft_loss} (\ie the sigmoid function) acts as a soft indicator for a hard negative triplet with $\Delta(h_n) > 0$. Similar to $\mathcal{L}_{\text{hard}}$, we propagate the loss gradient only through the nominator of \eqref{eq:soft_loss}.

The loss function $\mathcal{L}_{\text{soft}}$ encodes a soft triplet loss with a \textit{soft online hard example mining} strategy and requires only the vehicle's pose as ground truth. The gradient at the output of the ground and aerial networks $N_G$ and $N_A$ is sparse and dense respectively, but accumulated over multiple images for $N_G$.

\subsection{Tracking}

The method presented in the previous sections provides a means for registering aerial and ground data with small translational and rotational offsets, \ie with a limited number of hypotheses to test. The approach aims to achieve an upper bound on the long-term drift of a given tracking method, rather than to replace the tracking method itself.

We choose a simple tracking method based on an Extended Kalman Filter (EKF) with the constant turn-rate and acceleration (CTRA) motion model \cite{svensson2019derivation} that continuously integrates measurements of an inertial measurement unit (IMU). The EKF keeps track of the current vehicle state and corresponding state uncertainty. To demonstrate the effectiveness of our registration, we use only the turn-rate and acceleration of the IMU which on its own results in large long-term drift. The acceleration term is integrated twice to produce position values, such that small acceleration noise leads to large translational noise over time. We use our registration method to continuously align the trajectory with aerial images such that the drift is reduced to within the registration error of the method.

\subsection{Calibration of Confidence Scores}

The EKF takes as input the IMU measurement and a pose observation $z_{xy\phi} \in \mathbb{R}^3$ and observation uncertainty \mbox{$R_{xy\phi} \in \mathbb{R}^{3 \times 3}$} representing the network prediction with position $x$, $y$ and bearing $\phi$.

Determining $z_{xy\phi}$ and $R_{xy\phi}$ from the network output is \mbox{non-trivial} since the confidence scores predicted by the model are uncalibrated and do not represent the actual likelihoods of the underlying hypotheses. The confidence scores are further not guaranteed to follow a normal distribution which is required by the EKF.

We first transform all $s^{(h)}$ from $[-1, 1]$ into a proper likelihood function $L(h)$ as shown in \eqref{eq:s_to_p}.
\begin{equation}
	\label{eq:s_to_p}
	L(h) := \frac{s^{(h)} + 1}{\sum_{h' \in \mathcal{H}} (s^{(h')} + 1)}
\end{equation}

The observation $z_{xy\phi}$ is determined via maximum a posteriori estimation (MAP) using the likelihood $L$ and the prior probabilities $P_{\text{prior}}$ provided by the EKF.

The covariance $R_{xy\phi}$ cannot directly be estimated from the confidence scores which are uncalibrated and generally do not follow a normal distribution. Instead, we map $L$ into a more suitable representation $P = F(L)$ with a transformation $F$, compute its covariance \mbox{$\text{cov}(P)$} in the transformed space and apply the inverse transformation $F^{-1}$ to arrive at the actual covariance $R_{xy\phi}$ as shown in \eqref{eq:calibration}.
\begin{equation}
	\label{eq:calibration}
	R_{xy\phi} = F^{-1}(\text{cov}(F(L)))
\end{equation}
\begin{equation}
	\text{cov}(P) = \sum_{h \in \mathcal{H}} P(h)(h_{xy\phi} - z_{xy\phi}) \tran (h_{xy\phi} - z_{xy\phi})
\end{equation}

We define $F(L)$ as the posterior probability distribution over the set of hypotheses given the likelihood $L$ and prior probabilities $P_{\text{prior}}$. The forward transform $F$ thus multiplies $L$ and $P_{\text{prior}}$ and acts as a soft windowing function around the prior state estimate. Since this also decreases the estimated covariance in accordance with $P_{\text{prior}}$, the inverse transform $F^{-1}$ removes this effect and arrives at an undistorted covariance estimate $R_{xy\phi}$.

The choice of the transformation function $F$ is motivated by the fact that it is homomorphic when the network output already follows a normal distribution, \ie $F^{-1}(\text{cov}(F(L))) = \text{cov}(L)$. If the confidence scores do not follow a normal distribution, the transformation $F$ still allows capturing the curvature of the correlation volume in a soft window around the prior state estimate and convert it into a better calibrated normal distribution.

We discard unreliable registration results based on an empirical heuristic using the translational variance of the hypotheses, the confidence score of the predicted hypothesis and the mahalanobis distance of the predicted hypothesis to the prior probability distribution. In that case, only the IMU data is processed by the EKF.

\section{EVALUATION}

\subsection{Data}

To train and test our method, a sufficient number of geo-registered ground trajectories are required that contain both lidar point clouds and cameras ideally covering 360° around the vehicle. While many datasets for autonomous driving contain large numbers of ground samples, the trajectories often follow repeated routes with short distances and thus cover only few aerial images. Therefore, we choose four datasets for our training split (\ie Lyft L5 \cite{houston2020one}, Nuscenes \cite{caesar2020nuscenes}, Pandaset \cite{xiao2021pandaset}, Ford AV Dataset \cite{agarwal2020ford}), one dataset for validating the training (\ie Argoverse \cite{chang2019argoverse}) and report test results on two datasets (\ie KITTI odometry \cite{Geiger2013IJRR} and KITTI-360 \cite{liao2021kitti}). We use GoogleMaps \cite{googlemaps} and BingMaps \cite{bingmaps} to acquire orthophotos of the test split and the training/validation splits, respectively. Aerial and ground data are thus captured up to several years apart.

We train on datasets covering urban and sub-urban areas in the United States and test on datasets of an area around Karlsruhe, Germany. We thus demonstrate that our model is able to generalize over a large domain gap caused by 1.~the diverging appearance of regions in the United States and Germany, 2.~different capture modalities of the orthophotos and 3.~different capture modalities of the ground-based data.

We gather aerial images with a size of $512 \times 512$ pixels and a pixel resolution of $q = 0.2 \frac{\si{\m}}{\text{px}}$. We resize all ground images to a minimum size of $320 \times 240$ pixels, as we find this resolution to be sufficient for the registration task while enabling inference with multiple images on a single graphics processing unit (GPU). Since the locations of the ground data are distributed unevenly over the aerial images, we select ground samples during the training according to a uniform distribution over disjoint aerial cells of size $75 \si{\m} \times 75 \si{\m}$.

While modern autonomous driving datasets (\eg \cite{houston2020one,caesar2020nuscenes,xiao2021pandaset,agarwal2020ford,chang2019argoverse}) provide cameras with full 360° field-of-view, older datasets such as KITTI have only a front-facing stereo camera. Since our method utilizes lidar points that are covered by the viewing frustrum of at least one on-board camera, the older KITTI dataset cannot utilize its full potential. Despite these unfavorable conditions, our quantitative evaluation shows that our method outperforms other state-of-the-art geo-tracking appraoches on the KITTI dataset. We demonstrate our main results on the KITTI-360 dataset which contains a front-facing stereo camera and two lateral-facing fisheye cameras.

\subsection{Implementation}

We choose the ConvNeXt-T architecture \cite{liu2022convnet} as encoder and the decoder of UperNet \cite{xiao2018unified} for both the aerial and ground feature networks. The encoder is pre-trained on ImageNet \cite{deng2009imagenet}, and the entire encoder-decoder network is pre-trained on ADE20K \cite{zhou2019semantic}. The last layer is replaced with a $1 \times 1$ convolution to predict feature vectors with an embedding dimension of $c = 32$.

We train the model for 30 epochs with the RectifiedAdam optimizer \cite{liu2019variance} and a batch size of $1$. An epoch consists of a single ground sample chosen randomly per $75 \si{\m} \times 75 \si{\m}$ cell. We replace all BatchNorm (BN) layers \cite{ioffe2015batch} in UperNet with LayerNorm (LN) layers \cite{ba2016layer} since BN layers perform worse with small batch sizes \cite{wu2018group}. We employ a cosine decaying schedule with an initial learning rate of $3 \cdot 10^{-4}$. Generalization is improved via a decoupled weight decay \cite{loshchilov2017decoupled} of $1 \cdot 10^{-3}$, a layer decay \cite{bao2021beit} of 0.8 and an exponential moving average \cite{polyak1992acceleration} over the model weights with decay rate 0.999. The loss function is parametrized with a margin $m=0.1$ and a temperature $\delta=0.1$.

For data augmentation, we randomly rotate and flip the aerial image and lidar point clouds, and apply a color augmentation scheme to the aerial and ground camera images. We randomly shift the center of the aerial image by $0.4 \cdot h_A$ to prevent the network from simply learning a centering bias on the image. We choose the threshold $\alpha$ that discards unreliable hypotheses (\cf \secref{sec:feature_map_alignment}) as $\alpha = 0.3$ during training and $\alpha = 0.6$ during inference.

For the tracker we choose a total of $|\mathcal{H_R}| = 21$ possible rotations sampled from $-8$ to $+8$ degrees with the sampling density increasing for values closer to $0$.

\subsection{Results}

The experimental results on KITTI are summarized in \tabref{tab:kitti_ape}. We measure the performance as the mean APE of the two-dimensional poses in meters. Our method is able to track 8 of 10 scenes successfully and achieves state-of-the-art results on 7 scenes. Due to the limited camera field-of-view of KITTI, it loses track in two scenes at some point during the trajectory and is not able to recover afterwards. Only the method of Brubaker \etal \cite{brubaker2013lost} is able to successfully track the same number of sequences, but suffers from significantly larger APE.

The results on KITTI-360 are summarized in \tabref{tab:kitti360_ape}. Due to the larger field-of-view of the on-board cameras our method successfully tracks all sequences and achieves a mean APE of $0.94 \si{\m}$. Without the continuous alignment of our registration method the purely IMU-based tracker results in large long-term drift and an average APE of over $300 \si{\m}$.

\begin{figure}[t]
	\centering
	\vspace{2mm}
	\includegraphics[width=75mm]{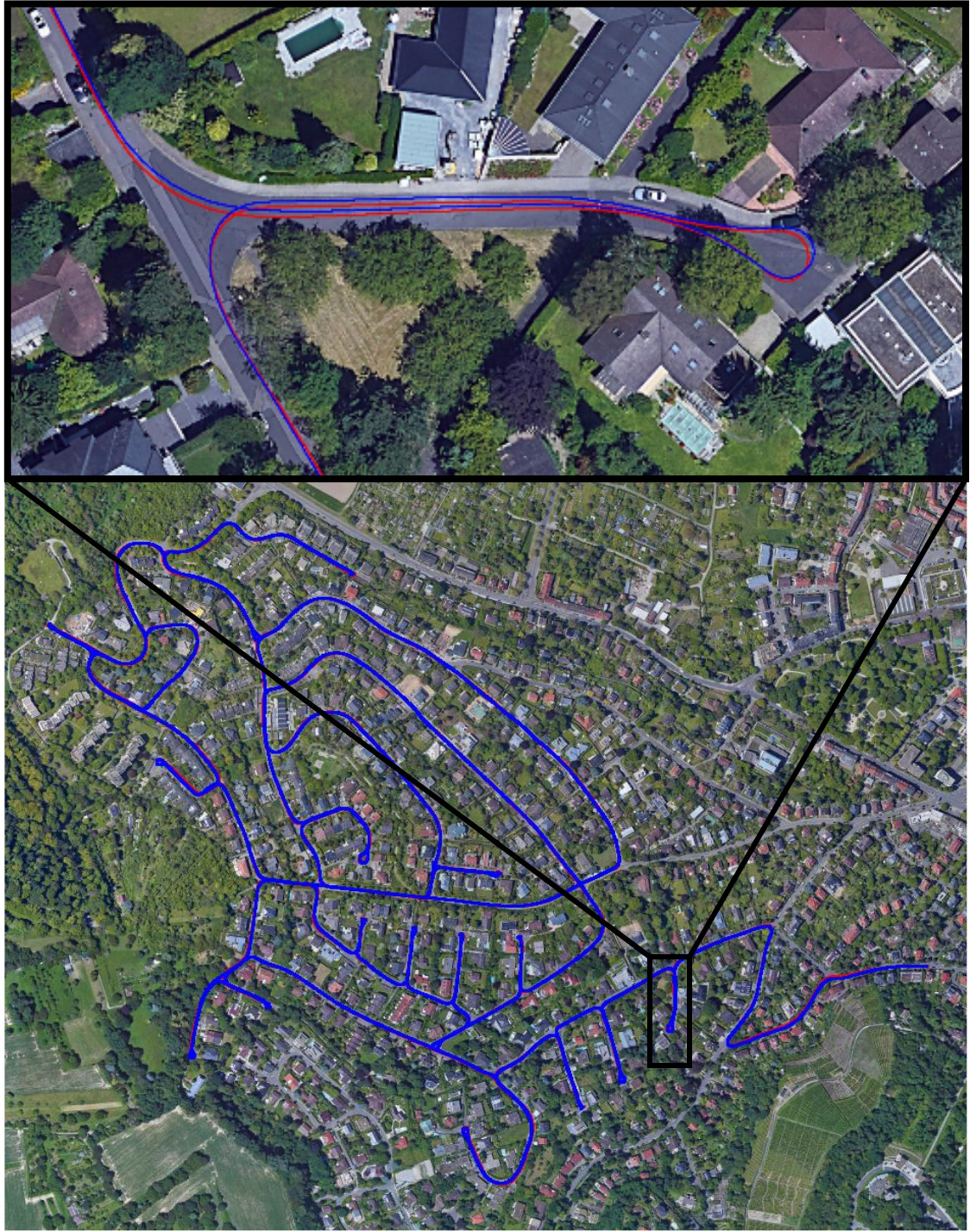}
	\caption{Scene 02 of the KITTI-360 dataset \cite{liao2021kitti} with the \mbox{ground truth} trajectory in red and the predicted trajectory in blue. Maps data \cite{googlemaps}: Google ©2022 Stadt Karlsruhe VLW, GeoBasis-DE/BKG (©2009).}
	\label{fig:trajectory}\vspace{-7mm}
\end{figure}

\figref{fig:trajectory} shows an overlay of the trajectory of scene 02 with aerial images of the area. The predicted and ground truth positions differ by $0.94\si{\m}$ on average and stay within an upper bound of $4\si{\m}$ to each other over the entire trajectory.

\begin{table*}[h]
	\centering
	\vspace{1.5mm}
	\caption{Absolute Position Error in meters on KITTI Odometry scenes. Results marked with parentheses correspond to a subset of the scene. Lower bounds are given for methods that did not report exact results. Missing cells indicate scenes that are not evaluated or not tracked successfully. Tang \etal \cite{tang2020rsl,tang2021self,Tang2021GetTT} show results for $x$ and $y$ dimensions separately - we report $\sqrt{x^2 + y^2}$. For each method, the best reported results are shown. Scene 03 is omitted since the GNSS ground truth is not available at the time of publication.}
	\vspace{-1mm}
	\label{tab:kitti_ape}
	\begin{tabular}{c|c|c|c|c|c|c|c|c|c|c}
		Method & 00 & 01 & 02& 04 & 05 & 06 & 07 & 08 & 09 & 10 \\
		\hline
		\rule{0pt}{2.1ex}Brubaker \etal \cite{brubaker2013lost} & 2.1 & 3.8 & 4.1 & -- & 2.6 & -- & 1.8 & \textbf{2.4} & \textbf{4.2} & 3.9\\
		Floros \etal \cite{floros2013openstreetslam} & $>10$ & -- & $>20$ & -- & -- & -- & -- & -- & -- & --\\
		Yan \etal \cite{yan2019global} & $>10$ & -- & -- & -- & $>10$ & $>10$ & $>10$ & -- & $>10$ & $>10$\\
		Miller \etal \cite{miller2021any} & \textbf{2.0} & -- & 9.1 & -- & -- & -- & -- & -- & 7.2 & --\\
		Tang \etal \cite{tang2020rsl,tang2021self,Tang2021GetTT} & -- & -- & (3.7) & -- & -- & -- & -- & -- & -- & --\\
		Zhu \etal \cite{zhu2020agcv} & -- & -- & -- & -- & -- & -- & -- & 4.45 & -- & --\\
		\textbf{Ours} & -- & \textbf{2.53}& \textbf{1.42} & \textbf{0.66} & \textbf{0.77} & \textbf{0.57} & \textbf{0.85} & 2.51 & -- &\textbf{0.96}\\
	\end{tabular}
\end{table*}

\begin{table*}[h]
	\centering
	\vspace{-1mm}
	\caption{Absolute Position Error in meters on KITTI-360 scenes}
	\label{tab:kitti360_ape}
	\hspace{-1mm}\begin{tabular}{c|c|c|c|c|c|c|c|c|c||c}
		Method & 00 & 02 & 03 & 04 & 05 & 06 & 07 & 09 & 10 & Mean\\
		\hline
		\rule{0pt}{2.2ex}\textbf{Ours} & 0.70 & 0.94 & 0.67 & 0.95 & 0.75 & 1.16 & 0.99 & 0.75 & 2.16 & 0.94\\
	\end{tabular}
	\vspace{-5mm}
\end{table*}

\section{CONCLUSION}

We present a novel geo-tracking method that allows tracking a vehicle on aerial images with meter-level accuracy. We perform evaluation on the \mbox{KITTI-360} and KITTI odometry datasets and report state-of-the-art results on the \mbox{geo-tracking} task. Our method is robust to changes in the environment and requires only geo-poses as ground truth. By training and testing on entirely different datasets we demonstrate its strong generalization. In the future, we will investigate the potential of the approach for change detection \wrt aerial imagery, and for identifying the misalignment of orthophotos \wrt \mbox{high-accuracy} GNSS signals.

\bibliographystyle{IEEEtran}
\bibliography{IEEEabrv,mybibfile}

\end{document}